\pdfoutput=1

\documentclass[11pt]{article}

\usepackage{acl}

\usepackage{times}
\usepackage{latexsym}

\usepackage[T1]{fontenc}

\usepackage[utf8]{inputenc}

\usepackage{microtype}

\usepackage{inconsolata}
\usepackage{microtype}
\usepackage{amsmath}
\usepackage{amssymb}
\usepackage{multirow}
\usepackage{booktabs}
\usepackage{bbm}
\usepackage{graphicx}
\usepackage{pifont}
\usepackage{caption}
\usepackage{subcaption}

\usepackage{xparse}
\NewDocumentCommand{\heng}
{ mO{} }{\textcolor{red}{\textsuperscript{\textit{Heng}}\textsf{\textbf{\small[#1]}}}}
\NewDocumentCommand{\jana}
{ mO{} }{\textcolor{red}{\textsuperscript{\textit{Jana}}\textsf{\textbf{\small[#1]}}}}

\NewDocumentCommand{\yubin}
{ mO{} }{\textcolor{red}{\textsuperscript{\textit{By Yubin}}\textsf{\textbf{\small[#1]}}}}
\NewDocumentCommand{\zx}
{ mO{} }{\textcolor{red}{\textsuperscript{\textit{By Ziang}}\textsf{\textbf{\small[#1]}}}}

\newcommand{\cmark}{\ding{51}}%
\newcommand{\xmark}{\ding{55}}%

\definecolor{entity}{RGB}{182,214,252}
\definecolor{relation}{RGB}{243,191,189}

\title{What should I Ask: A Knowledge-driven Approach for Follow-up Questions Generation in Conversational Surveys}

\author{Yubin Ge\thanks{\ \ denotes equal contribution.}\textsuperscript{$*1$}, Ziang Xiao\textsuperscript{$*1$}, Jana Diesner\textsuperscript{$1$}, Heng Ji\textsuperscript{$1$}, Karrie Karahalios\textsuperscript{$1$}, Hari Sundaram\textsuperscript{$1$} \\
  \textsuperscript{$1$}University of Illinois Urbana-Champaign, USA \\
  \texttt{\{yubinge2,zxiao5,jdiesner,hengji,kkarahal,hs1\}@illinois.edu} \\}

\begin{document}
\maketitle
\begin{abstract}
Generating follow-up questions on the fly could significantly improve conversational survey quality and user experiences by enabling a more dynamic and personalized survey structure. In this paper, we proposed a novel task for knowledge-driven follow-up question generation in conversational surveys. We constructed a new human-annotated dataset of human-written follow-up questions with dialogue history and labeled knowledge in the context of conversational surveys. Along with the dataset, we designed and validated a set of reference-free Gricean-inspired \cite{grice1975logic} evaluation metrics to systematically evaluate the quality of generated follow-up questions. We then propose a two-staged knowledge-driven model for the task, which generates informative and coherent follow-up questions by using knowledge to steer the generation process. The experiments demonstrate that compared to GPT-based baseline models, our two-staged model generates more informative, coherent, and clear follow-up questions. 

\end{abstract}

\section{Introduction}
A conversational survey collects information from people through an open-ended conversation where an agent asks questions, interprets responses, and probes answers \citep{xiao2020tell,gobo2011back}. Compared to structured form-based surveys, a conversational survey enables a more dynamic survey structure and personalized experience through follow-up questions \cite{xiao2020if}. Although a good follow-up question could probe and prompt more information based on people's responses, the current practice of delivering follow-up questions in conversational surveys is largely rule-based \cite{grudin2019chatbots}. In this study, we explored an automatic way of generating follow-up questions for conversational surveys.

\begin{figure}  
\centering
\includegraphics[width=0.8\linewidth]{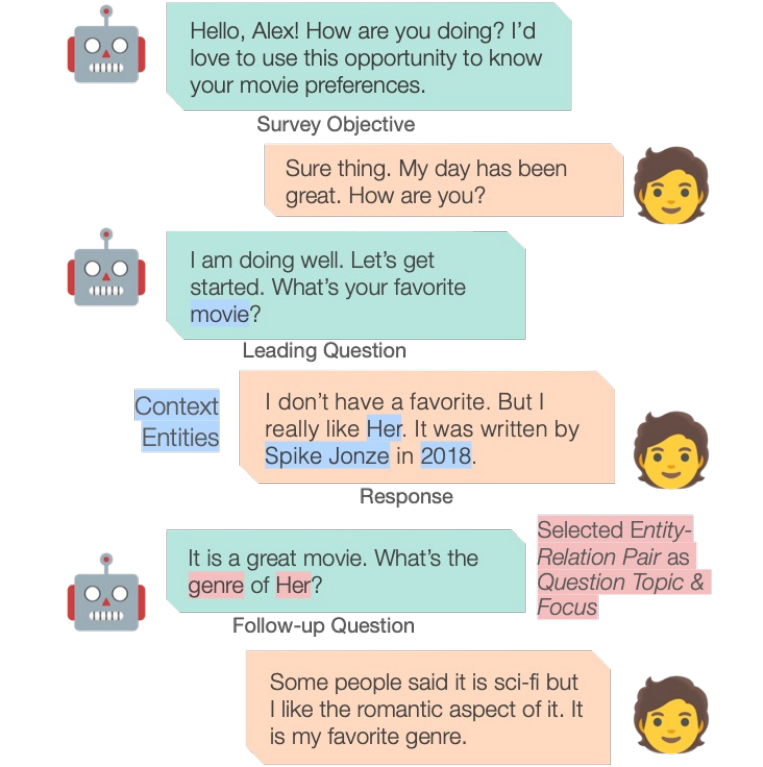}
    \caption{A conversational survey snippet that aims to understand people's movie preferences where the agent generates a follow-up question based on the dialogue history and a selected entity-relation pair.} 
    \label{fig:opening}
\vspace{-5mm}
\end{figure}

Although the conversational survey could be used for various objectives, e.g., personal experiences, public opinion, etc.,  on different topics, e.g., daily activities, domain knowledge, etc., a single conversational survey often focuses on one survey objective. Therefore, a good follow-up question should be concise and relevant to the survey objective and should be phrased in a clear manner. It requires the system to understand the context deeply, adhere to the survey objective, and generate questions that are easy to understand. 

To enable the automatic generation of follow-up questions in an open-domain conversational survey, we aim to tackle three specific challenges. First, there is no dataset for follow-up question generation in the context of open-domain conversational surveys. Related datasets are small or with a very specific focus, such as job interviews \citep{sb2020automatic} or graduate school admission interviews \citep{su2018follow,su2019follow}. Additionally, no prior dataset considers the background knowledge beyond the dialogue history, which limits the model's ability to deeply understand the survey objective and context.

Second, existing methods for follow-up question generation are either template filling \citep{su2019follow,Inoue2020JobIA} or seq2seq \citep{su2018follow,wang2018learning,sb2020automatic}. However, both methods lead to unsatisfying results. Template filling limits the diversity of question types and often fails to personalize based on the participant's response, especially in dynamic and open-ended conversational surveys. As for standard seq2seq methods, they cannot generate questions that adhere to the overall survey objective and are relevant to the context. How to improve the diversity of generated follow-up questions while not limited by templates while utilizing relevant background knowledge remains an unresolved problem.

Third, no established metrics can effectively evaluate the generated follow-up questions for conversational surveys. The same dialogue history can inspire various valid follow-up questions, and the same question can be phrased differently. Hence, common text generation metrics depending on only one ground truth usually underestimate question quality, and human evaluation is hard to scale and compare \cite{dinan2018wizard,sb2020automatic,xiao2023evaluating}.

In this study, we proposed a task of knowledge-driven follow-up question generation in conversational surveys, as shown in Fig. \ref{fig:opening}. To explore its feasibility and effectiveness, we collected a human-annotated dataset including background knowledge, proposed a two-staged knowledge-driven baseline model, and designed a set of reference-free Gricean-inspired evaluation metrics. 

Specifically, we first collected a dataset with knowledge annotation and a human-written follow-up question based on the dialogue history and background knowledge. The follow-up question aimed to collect relevant and valuable information in a coherent and clear manner. To demonstrate feasibility, we proposed a baseline model that leverages knowledge to steer the generative model. We imitated the human question-generation process by using a knowledge selector to identify the question topic and focus \cite{duan2008searching} based on the dialogue history and a background knowledge base. We then combined the dialogue history with the selected question topic and focus as a prompt for a generative model to generate follow-up questions that could lead to valuable and diverse information that contributes to the survey objective. To systematically evaluate the quality of generated follow-up questions, we designed a new set of reference-free evaluation metrics, \textit{Gricean Scores}, based on Gricean Maxims \citep{grice1975logic}. Gricean Maxims is a collection of communication principles to which both speaker and listener should adhere to engage in effective communication. Our \textit{Gricean Scores} (see Sec. \ref{griceanscores}) measure the quality of a follow-up question based on the following aspects: \textit{Relevance}, \textit{Informativeness}, \textit{Truthfulness}, \textit{Clarity}, and \textit{Coherence}. \textit{Gricean Scores} align with human evaluation well and can provide more insights than traditional metrics. 

Our contributions are as follows: 

$\bullet$ A new problem and a dataset for follow-up question generation in conversational surveys, which has background annotation and human-written follow-up questions.

$\bullet$ An effective baseline model that leverages knowledge as a control to generate informative, coherent, and clear follow-up questions. 

$\bullet$ A set of reference-free evaluation metrics based on Gricean Maxims, i.e., \textit{Gricean Scores}, that evaluate the quality of open-ended text from various perspectives.

\section{Dataset and Annotation}

Three principles guide our dataset construction: first, the topic needs to be diverse as conversational surveys can cover many topics. Second, the follow-up question should adhere to the conversation context, and retrieve novel and specific information from the respondent. Third, the knowledge should be explicit, especially salient knowledge that guides the follow-up question generation.

\subsection{Question-Answer Pairs Construction}

Our dataset is based on OpenDialKG \citep{moon2019opendialkg}, which contains open-domain conversations between two human agents about a given topic, such as movies, books, sports, and music. Those topics often appear in conversational surveys to build rapport and collect various responses. Each utterance in the original dataset has been manually annotated with entities from Freebase \citep{bast2014easy}. We first extracted question-answer pairs from OpenDialKG and then manually selected those that have a clear information goal, e.g., ``how are you doing?'' will not be included.

\subsection{Follow-up Questions Annotation}
We invited workers on Amazon Mechanical Turk to create follow-up questions for each question-answer pair. We first presented a dialogue history, e.g., a question-answer pair, with the overarching survey objective and instructed people to imagine themselves as an interviewer who aims to collect informative and truthful information that contributes to the overall survey objective from their interviewees through a follow-up question. We derive the survey objective based on the topic of the extracted Q \& A pair, e.g, [topic] preference. To facilitate the creation of high-quality follow-up questions, we first asked people to select the most interesting and meaningful topic that they wanted to ask about in the follow-up question from mentioned knowledge entities in the dialogue history. Then, they needed to specify the relation of the selected entity as the follow-up question focus. As the last step, we instructed people to write a follow-up question in a clear and coherent manner based on the above-mentioned criteria (Sec. \ref{appendix:statistics}). We manually went through all annotations and made minor edits to ensure quality. We compensated our workers at the rate of \$12 per hour.

\section{Models}
\begin{figure*}
    \centering
    \includegraphics[width=0.8\linewidth]{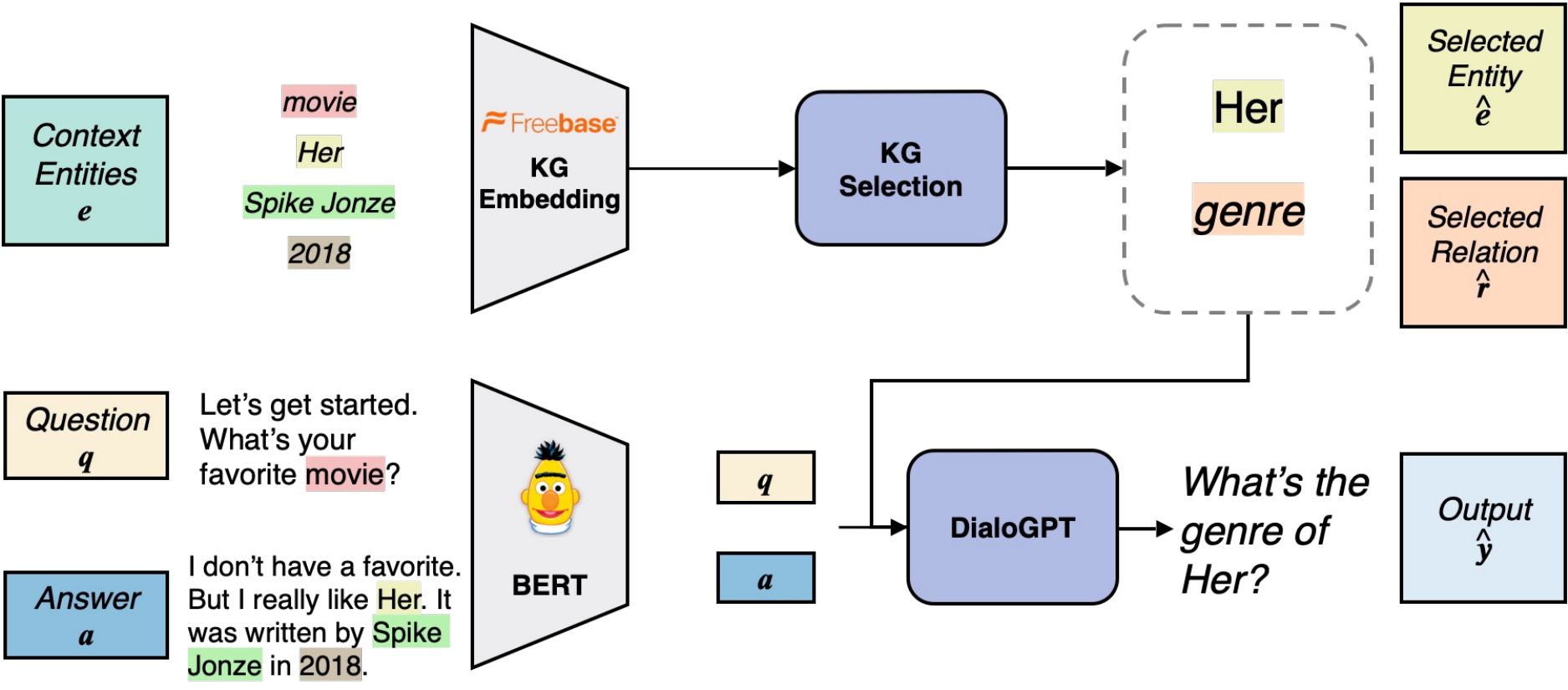}
    \caption{The overall architecture of the proposed framework.} 
    \label{fig:framework}
\vspace{-5mm}
\end{figure*}
\subsection{Overview}

We formalized the task of follow-up question generation as follows: given a dialogue history $X_i$ of the last two turns, consisting of a QA pair $(Q_i, A_i)$, with a set of context entities $\mathcal{E}_i = \{e_1, e_2, ..., e_n\}$ (e.g., \colorbox{entity}{Her}, \colorbox{entity}{Spike Jonze} highlighted in Fig. \ref{fig:opening}) and their relations $\mathcal{R}$ (e.g., the relation \colorbox{relation}{genre} for the entity \colorbox{entity}{Her} highlighted in Fig. \ref{fig:opening}), a system needs to generate a follow-up question $Y_i$ that is specific to the selected entity-relation pair $(e_j, r_k)$. Here $e_j \in \mathcal{E}_i$, $r_k \in \mathcal{R}(e_j)$, and we denote $\mathcal{R}(e_j)$ as the set containing all relations connected to the entity $e_j$.

Inspired by how human experts generate questions in semi-structured interviews \cite{wilson2013interview}, we propose a two-staged framework (as shown in Fig. \ref{fig:framework}) that (i) selects an entity-relation pair as the question topic and focus that conditioned on the dialogue history (Sec. \ref{sec:selection}), and then (ii) generates a follow-up question that is triggered by the selected entity-relation pair (Sec. \ref{sec:generation}).  

\subsection{Knowledge Selection}
\label{sec:selection}
Our knowledge selection model first encodes a dialogue history $X_i$, a QA pair $(Q_i, A_i)$, using BERT \citep{devlin2019bert}. We process the input text as a concatenation of the question $Q_i$ and its answer $A_i$, and insert a special tag {\tt [CLS]} at the beginning and another special tag {\tt [SEP]} to separate them. The final hidden state that corresponded to {\tt [CLS]} is used as the aggregated sequence representation of the dialogue history, which is denoted as $\mathbf{h}_i^{\text{qa}}$. Meanwhile, we encode entities and relations with pre-trained knowledge embedding (e.g., TransE \citep{NIPS2013_1cecc7a7}), and we represent them as $\mathbf{h}_j^{\text{e}}$ and $\mathbf{h}_k^{\text{r}}$ for the entity $e_j$ and the relation $r_k$.

We consider two methods to perform the knowledge selection: attention-based and MLP-based. The entity selection works in the same way as the relation selection. Here we only introduce the entity selection for simplicity. In the attention-based version, we regard the score of selecting the entity $e_j$ as the unnormalized dot-product attention score \citep{vaswani2017attention} between $\mathbf{h}_j^{\text{e}}$ and $\mathbf{h}_i^{\text{qa}}$:
\begin{align}
    & \mathbf{s}_i = \frac{\mathbf{q}_i\mathbf{K}_i^T}{\sqrt{d_k}},
\end{align}
where $\mathbf{q}_i$ is the query vector such that $\mathbf{q}_i = \mathbf{W}^Q \mathbf{h}_i^{\text{qa}}$, $\mathbf{K}_i$ is the key matrix which is a stack of key vectors $\mathbf{W}^K \mathbf{h}_j^{\text{e}}$ for all $e_j \in \mathcal{E}_i$, and $d_k$ is the dimension of queries and keys. As for the MLP-based version, we concatenate  $\mathbf{h}_i^{\text{qa}}$ and $\mathbf{h}_j^{\text{e}}$, then input it into an MLP to predict the score:
\begin{align}
    & s_{ij} = \text{MLP}([\mathbf{h}_i^{\text{qa}};\mathbf{h}_j^{\text{e}}]),
\end{align}
where $[.;.]$ denotes the concatenation operation.

Finally, we treat the selection as a binary classification for each candidate entity and set the objective as minimizing the binary cross entropy:
\begin{align}
    \mathcal{L}_{\text{ent}} = - &\frac{1}{N}\sum_{i=1}^N\sum_{j:e_j \in \mathcal{E}_i}[y_j^i\cdot \log \sigma(s_{ij}) \notag \\
    & + (1-y_j^i)\cdot\log(1-\sigma(s_{ij}))],
\end{align}
where $\sigma(.)$ is the sigmoid function, $N$ is the number of samples, $y_j^i$ is the ground truth label for the $j^{th}$ entity in the $i^{th}$ sample, and it equals $1$ if the corresponding entity is selected and is $0$ otherwise. During the training, we take the sum of $\mathcal{L}_{\text{ent}}$ and $\mathcal{L}_{\text{rel}}$ as the final objective function.

\subsection{Follow-up Question Generation}
\label{sec:generation}
We formulate follow-up question generation as a typical language modeling problem and adopt DialoGPT \citep{zhang-etal-2020-dialogpt} to solve it. 
 
We serialize the input into one sequence and then input it into DialoGPT to estimate the language model. Specifically, we first concatenate $Q_i$, $A_i$ and $Y_i$, and add {\tt [EOS]} at the end of each utterance such as $[Q_i, \text{{\tt EOS}}, A_i, \text{{\tt EOS}}, Y_i, \text{{\tt EOS}}]$. With the selected entity $e_i^*$ and relation $r_i^*$, we manually design a discrete template, such as "How to ask about", to combine $e_i^*$ and $r_i^*$ into a prompt $P_i$. The prompt is treated as an utterance, and is inserted between the dialogue history and the follow-up question to guide the generation. The final input can be represented as one sequence of tokens $[u_1, u_2, ..., u_m] = [Q_i, \text{{\tt EOS}}, A_i, \text{{\tt EOS}}, P_i, \text{{\tt EOS}}, Y_i, \text{{\tt EOS}}]$.  At each decoding step $t$, we compute the negative log-likelihood between the predicted distribution and the ground truth from the given response:
\begin{align}
    & \mathcal{L}_{\text{NLL}} = -\sum_t\log p(u_t |u_{<t})
\end{align}
Notably, during the inference time, we only input $[Q_i, \text{{\tt EOS}}, A_i, \text{{\tt EOS}}, P_i, \text{{\tt EOS}}]$ into DialoGPT and let it decode the response until reaching {\tt [EOS]} as the generated follow-up question.

\section{Gricean Scores}
\label{griceanscores}
To systematically assess the quality of generated follow-up questions, we proposed a set of reference-free, customizable and content-based metrics, \textit{Gricean Scores}, grounded on Gricean Maxims \citep{grice1975logic}. The Gricean Maxims is a collection of communication principles, to which both speaker and listener should adhere to engage in effective communication. Researchers have been using Gricean Maxims to evaluate both human-human conversations \cite{eskritt2008preschoolers,kleinke2010speaker} and human-agent conversations \cite{xiao2020tell,langevin2021heuristic}. In the context of conducting conversational surveys, a ``cooperative'' interviewer would obey all the maxims to form questions and probe quality responses. Those maxims include \textit{Quantity}, \textit{Quality}, \textit{Relation}, and \textit{Manner}.

We relied on the Gricean Maxims to define a set of reference-free metrics that quantitatively measure the quality of a follow-up question generated by our models. We measure the quality of a follow-up question and compute \textit{Gricean Scores} from five aspects: \textit{Relevance}, \textit{Informativeness}, \textit{Truthfulness}, \textit{Clarity}, and \textit{Coherence}. Our reference-free metrics allowed us to evaluate a follow-up question without relying on a single ground truth, making it more robust to open-ended text generation and real-world deployment. Also, the \textit{Relevance}, \textit{Informativeness}, and \textit{Truthfulness} are knowledge-driven, which makes our score more contextualized. The survey designers could bring a knowledge graph, customize one, or create one that best fits the survey objective and information value. 

\textbf{Notation.} We defined indexed sets $\hat{\mathcal{Q}}$, $\mathcal{T}$ such that a generated follow-up question $\hat{Q}_i \in \hat{\mathcal{Q}}$ corresponds to a target follow-up question $\mathcal{T}_i \in \mathcal{T}$. We referred the dialogue history of $\hat{Q}_i$ as its context $C_i$ and further represent $\hat{Q}_i$ as a sequence of tokens $[w_1, w_2, ..., w_N]$. We marked the recognized entity from $\hat{Q}_i$ as $\hat{e}_i$, and the set of context entities as $\mathcal{E}_i$.

\textbf{Relevance.} By the Gricean Maxim of relation, a high-quality follow-up question should be relevant to the prior discussion and adhere to the overall survey objective. Irrelevant follow-up questions elicit useless information and burden the analysis process. We measured \textit{Relevance} by checking if the recognized entity in the generated follow-up question is from the context entity set:

\begin{align}
    & \textbf{REL}(\hat{Q}_i) = \mathbbm{1}[\hat{e}_i \in \mathcal{E}_i] 
\end{align}

\textbf{Informativeness.} From the Gricean Maxim of Quantity, effective communication should be informative. In the context of a conversational survey, an agent should always ask questions that maximize the information gain in a participant's response. In a knowledge graph, an entity's out-degree centrality captures the outreach of other entities which indicates the potential information gain by probing this entity. We measured the \textit{Informativeness} of a generated follow-up question by the out-degree centrality \footnote{The centrality is normalized by dividing by the maximum possible degree in the graph. For multi-turn conversation, when calculating the out-degree centrality, some edges were removed if mentioned in previous conversation turns.} of the recognized entity in the question: 

\begin{align}
    & \textbf{INFO}(\hat{Q}_i) = \text{Centrality}(\hat{e}_i)
\end{align}

\textbf{Truthfulness.} According to the Gricean Maxim of Quality, a cooperative interlocutor should communicate truthfully. In our case, a high-quality follow-up question should always lead to truthful information. For example, a survey respondent would respond to a question asking for a book's release year but not a question about a book's director since the book does not have such an attribute. We measured \textit{Truthfulness} of a follow-up question from an entity-relation perspective where the question should ask a valid relation connected to the recognized entity based on the knowledge base \footnote{To align with the knowledge base we select for our dataset, we choose Freebase here.}. 

Specifically, we first trained a BERT-based relation prediction model (similar to the knowledge selection model introduced in Sec. \ref{sec:selection}) to predict which relation is contained in $\hat{Q}_i$, and we denoted the predicted relation as $\hat{r}_i$. Then we measured \textit{Truthfulness} of $\hat{Q}_i$ as checking whether $\hat{r}_i$ is connected to $\hat{e}_i$ in the knowledge base \textbf{KG}:
\begin{align}
    & \textbf{TRUTH}(\hat{Q}_i) = \mathbbm{1}[(\hat{e}_i, \hat{r}_i) \in \textbf{KG}] 
\end{align}

\textbf{Clarity.} The Gricean Maxim of Manner advocates that one should communicate in a clear and orderly manner, and therefore we require a follow-up question to avoid obscurity and ambiguity. We regarded \textit{Clarity} of a generated follow-up question as how well it fits in natural language, and we used an external powerful language model as an expert to provide the measurement. Hence, the perplexity of a generated follow-up question is computed under a pre-trained language model and we pick DialoGPT given it's pre-trained on dialogue data. Specifically, we calculate it as follows:
\begin{align}
    & \textbf{CLA}(\hat{Q}_i) = \text{exp}(-\frac{1}{N}\sum_j^N\log p_\theta (w_j|w_{<j}))
\end{align}

\textbf{Coherence.} A coherent conversation reflects another aspect of Manner as well. Following previous work \citep{bommasani2020intrinsic}, we evaluated the semantic \textit{Coherence} of generated follow-up questions to dialogue histories by predicting the probability of each generated question conditioned on the previous QA pair using the powerful language model, BERT, whose pre-training tasks include this same objective.
\begin{align}
    & \textbf{COH}(\hat{Q}_i) = \mathbbm{1}_{\text{BERT}}(\hat{Q}_i | C_i)
\end{align}

\begin{table*}
\small
\centering
\scalebox{0.95}{
\begin{tabular}{lcccc}
\toprule
 \multicolumn{1}{c}{\multirow{2}{*}{\bf Model}} & \multicolumn{1}{c}{\bf Entity Selection} & \multicolumn{3}{c}{\bf Relation Selection} \\ \cmidrule(r){2-2} \cmidrule(r){3-5} & R@1 & R@1 & R@3 & R@5 \\ \midrule 
\multicolumn{5}{l}{\textbf{Attention-based}} \\\midrule
KG Selector (w/ TransE) & 0.642 & 0.271 & 0.593 & 0.761 \\
\phantom{KG Selector} (w/ TransD) & 0.624 & 0.295 & 0.616 & 0.773 \\
\phantom{KG Selector} (w/ TransR) & 0.628 & 0.288 & 0.612 & 0.773 \\
\midrule
\multicolumn{5}{l}{\textbf{MLP-based}} \\\midrule
KG Selector (w/ TransE) & 0.647 & 0.283 & 0.605 & 0.768 \\
\phantom{KG Selector} (w/ TransD) & 0.648 & 0.288 & 0.613 & 0.763 \\
\phantom{KG Selector} (w/ TransR) & \textbf{0.654} & \textbf{0.302} & \textbf{0.620} & \textbf{0.776} \\ \bottomrule

\end{tabular}}
\caption{\label{selection_result} Experimental results of various methods on the knowledge selection task.}
\vspace{-0.2cm}
\end{table*}

\section{Experiments}
 We provide the details of experiments in this section and include implementation details in Sec. \ref{appendex:implementation}.

\subsection{Knowledge Selection}
We first assessed our model's prediction of the salient knowledge chosen by humans given the dialogue history. The result can inform us of the feasibility of explicitly separating knowledge selection from the full task and choosing the best model to use in our two-staged framework. Meanwhile, separating this task could increase the interpretability of the whole framework by explicitly identifying the knowledge entity and relation that steer the follow-up question generation. We compared the two proposed baseline models, attention-based and MLP-based, and examined how well they interact with different types of knowledge embedding, including TransE \citep{NIPS2013_1cecc7a7}, TransR \citep{lin2015learning} and TransD \citep{ji2015knowledge}. 

Tab. \ref{selection_result} shows the results of recall$@$k of the selection models. MLP-based models generally perform better than attention-based models for both selection tasks, and the MLP-based model with TransR reached the best performance with $0.654$ recall$@$1 score on entity selection and $0.302$ recall$@$1 score on relation selection. Based on this, we decided to use the best-performing model, MLP-based with TransR, in the whole framework.

\begin{table*}
\centering
\scalebox{0.95}{
\begin{tabular}{lccccc}
\toprule
 \bf Model & \textbf{REL}($\%$) $\uparrow$ & \textbf{INFO} $\uparrow$ & \textbf{TRUTH}($\%$) $\uparrow$ & \textbf{CLA} $\downarrow$ & \textbf{COH}($\%$) $\uparrow$ \\ \midrule 
KG-FQG (w/o knowledge)  & 60.06 & 0.47 & 41.63 & \bf 2.46 & \bf 99.07 \\
\phantom{KG-FQG} (w/ predicted knowledge)  & \bf 72.51 & 0.55 & \bf 67.93 & 2.92 & 98.55 \\
\phantom{KG-FQG} (w/ gold-standard knowledge)  & 69.42 & \bf 0.63 & 63.35 & 2.83 & 98.99 \\ \bottomrule

\end{tabular}}

\caption{\label{generation_result} Results of various methods on the follow-up question generation task. $\uparrow$ indicates the higher score the better, while $\downarrow$ means the lower score the better.}
\vspace{-0.2cm}
\end{table*}

\begin{table}
\small
\centering
\scalebox{0.85}{
\begin{tabular}{lccc}
\toprule
 \bf Model & \bf R-1 & \bf R-2 & \bf R-L \\ \midrule 
KG-FQG (w/o knowledge)  & 15.46 & 3.82 & 15.06  \\
\phantom{KG-FQG} (w/ predicted knowledge)  & 19.66 & 7.12 & 19.32 \\
\phantom{KG-FQG} (w/ gold-standard knowledge)  & \bf 34.98 & \bf 16.58 & \bf 34.00 \\ \bottomrule

\end{tabular}}
\caption{\label{rouge_result} ROUGE scores of various methods on the follow-up question generation task.}
\vspace{-0.5cm}
\end{table}

\subsection{Follow-up Question Generation With Knowledge}
We evaluated our model on the follow-up question generation with selected knowledge in two settings: given the gold-standard knowledge chosen by human annotators, or the knowledge predicted by the best selection model. Additionally, we compared them with the baseline model DialoGPT without adding external knowledge to evaluate the effectiveness of explicitly employing knowledge. 

We first used \textit{Gricean Scores} to perform the evaluation and analyze the results using ANOVA tests with Tukey methods \citep{miller1997beyond}. As shown in Tab. \ref{generation_result}, compared to the baseline model, our proposed knowledge-driven models significantly improved the quality of generated follow-up questions in terms of \textit{Relevance} ($F(2,3009) = 19.41, p<0.01^{**}$) and \textit{Truthfulness} ($F(2,3009) = 85.67, p<0.01^{**}$). This suggests that integrating selected knowledge is a valid approach to leverage knowledge to guide the generation. For \textit{Clarity} we found the baseline model performed the best ($F(2,3009) = 15.59, p<0.01^{**}$). 
As for \textit{Informativeness} and \textit{Coherence}, three models all achieved good performance with insignificant difference, and we think it's due to the powerful backbone, DialoGPT, we chose. We found the difference between our model with predicted knowledge and gold-standard knowledge for all five dimensions is not statistically significant. This shows that even if the chosen knowledge by the knowledge selection model is different from the ones by human annotators, our generation model can generate appropriate follow-up questions, and demonstrated the applicability of our proposed method for the open-ended scenario. 
By examining the outputs, we noticed that including external knowledge through prompts can reduce hallucinations - the fabrication of untruthful information \citep{maynez-etal-2020-faithfulness}. For example, when a dialogue history mentions "Oh, I loved Tom Arnold in \textit{Undercover Blues}", the baseline model that does not involve knowledge generates "Who else starred in \textit{Underpants Blues}", while our model is capable of generating the movie name correctly. This indicates that adding external knowledge through prompts can be a potential solution to entity-level hallucination in text generation.

We further adopt the common metric ROUGE \citep{lin2004rouge} and report the results in Tab. \ref{rouge_result}.  Our experiments again showed that adding knowledge is an effective way to involve knowledge in follow-up question generation, as the two models with knowledge outperform their counterpart, DialoGPT without knowledge. However, we can see there is still a gap between the model with predicted knowledge and gold-standard knowledge, and we think the reason comes from the different knowledge selections, which results in generating totally different follow-up questions. For example, asking about an actor and a movie can be different, but both valid follow-up questions. However, the only-one ground truth for one test instance limits the consideration of other possibilities, and this may underestimate models' performance in open-ended text generation. By comparing to the results above, this shows the superiority of our \textit{Gricean Scores} which evaluates multiple aspects of generated text without relying on ground truth text. The reference free nature of the \textit{Gricean Scores} opens its way to the evaluation of other text generation tasks.

\subsection{Expert Evaluation}
Complementing the result with our objective metrics, we randomly sampled 50 instances from the generated follow-up questions and ground truth texts to perform an expert evaluation. We first discussed a codebook based on four dimensions of Gricean Maxim on a 0 (poor) to 2 (excellent) scale. Then two expert annotators individually rated instances blind to the generation methods. Krippendorff’s alpha ranged from 0.80 to 0.92 for each set of coding. We averaged their scores (Tab. \ref{human_result}) and performed ANOVA tests with Tukey methods to analyze the results. Overall, the results aligned with our proposed \textit{Gricean Scores}, where the knowledge-driven methods outperform the baseline model, DialoGPT, in all dimensions except \textit{Manner}. The Tukey post-hoc tests showed the differences between Ground Truth and our knowledge-driven methods are not statistically significant in all four dimensions, indicating the effectiveness of our approach. In addition, our \textit{Gricean Scores} correlates well with our expert evaluation, Quantity: 0.16, Quality = 0.37*, Relation = 0.24*, and Manner = 0.19 \footnote{* indicates the correlation is significant per Pearson's test}.

\begin{table*}
\centering
\scalebox{0.95}{
\begin{tabular}{lcccc}
\toprule
 \bf Model & \textbf{Quantity} & \textbf{Quality} & \textbf{Relation} & \textbf{Manner} \\ \midrule 
DialoGPT  & $1.41$ & $1.52$ & $1.40$ & $1.62$  \\
Ground Truth  & $1.82^{**}$ & $1.95^{**}$ & $1.82^{**}$ & $1.84$  \\
KG-FQG (w/ predicted knowledge)  & $1.74^{*}$ & $1.86^{**}$ & $1.66$ & $1.84$  \\
\phantom{KG-FQG} (w/ gold-standard knowledge)  & $1.80^{**}$ & 1.74 & $1.77^{**}$ & $1.75$ \\ \bottomrule

\end{tabular}}

\caption{\label{human_result} Human Evaluation results of various methods on the follow-up question generation task on Likert Scales from 0-2. The $*$ indicates the difference against DialoGPT is statistically significant. $^* p< 0.05$. $^{**} p < 0.01$. }
\vspace{-0.2cm}
\end{table*}

\section{Error Analysis}
\begin{table}
\small
\centering
\scalebox{0.95}{
\begin{tabular}{ll}
\toprule
Dialogue & -- Was Two by Two released in 2016? \\
History & -- It was, the same year \underline{When I'm} \\ & \underline{Gone} was released, which is \\ & another similar book. \\ \midrule
Predicted Entity: & \textit{When I'm Gone} \\
Predicted Relation: & \textit{release\_year} \\
KG-FQG: & -- When was "When I'm gone" \\ & released? \\ \midrule
Dialogue & -- Do you know \underline{The Runaway Jury}? \\
History & -- The Runaway Jury is written by \\ & John Grisham, with a genre of \\ &  Suspense. \\ \midrule
Predicted Entity: & \textit{The Runaway Jury} \\
Predicted Relation: & \textit{subject} \\
KG-FQG: & -- What is the subject of the movie? \\ \bottomrule

\end{tabular}}
\caption{\label{error} Generated examples from KG-FQG with predicted knowledge for error analysis.}
\vspace{-0.2cm}
\end{table}

Tab. \ref{error} shows several examples that our model with predicted knowledge fails. In the first example, we can observe that our model picks an appropriate mentioned entity, the book \textit{When I'm gone}, to formulate its follow-up question but asks the relation \textit{release\_year} which has been covered in the dialogue history. This implies that our current model sometimes may not pay attention to the dialogue history when selecting a relation for asking. We believe this issue might be solved if we add additional constraints for relation prediction and only allow the knowledge selection model to select unmentioned relations.

Another kind of error can be seen in the second example. The dialogue history mentions \textit{The Runaway Jury}, which is a novel, but our model misunderstands it as a movie even if the phrase "is written by" in the dialogue history indicates it is a book. We notice that such a problem usually happens when different entities share the same name, such as an adapted movie and its original book. Therefore, it is necessary to perform named entity disambiguation during the preprocessing, and exploring better knowledge embedding may be potentially useful in alleviating this problem as well.

\section{Related Work}
\textbf{Follow-up Question Generation} Previous attempts have explored this task in more specific domains and followed the common pipeline in various text generation tasks \cite{ge2021baco,ge2023detection}. For example, several attempts focused on graduate school admission interviews \cite{su2018follow,su2019follow} and built a small corpus by simulating interviews between participants. They have tried template filling-based and sentence retrieval-based methods for generating follow-up questions. Additionally, job interview is another main domain of the task. To this end, \citet{sb2020automatic} adopted pre-trained GPT-2 and fine-tuned it on their own collected corpus. By contrast, \citet{Inoue2020JobIA} pre-defined a set of follow-up questions under different categories and turned the generation task as a question selection problem. Another line of related research lies in question generation in conversational systems. For example, \citet{wang2018learning} devised typed decoders to model a type distribution over \textit{interrogatives}, \textit{topic words} and \textit{ordinary words}, and used it to modulate the final generation distribution. By contrast, we focused on the scenario of conversational surveys where the information collection goal is more diverse. We also exploit a knowledge base to guide the generative model to generate more focused and informative questions. 

\textbf{Dialogue Evaluation Metrics}
Researchers have shown that standard automatic language evaluation metrics (e.g., BLEU, METEOR) are ineffective for dialogue evaluation \citep{liu2016not, deriu2021survey, yeh2021comprehensive}. Hence, recent research has proposed various automatic metrics specifically for dialogue generation, which can be divided into two categories, reference-required and reference-free, based on whether a reference utetrance is required. 

Among reference-required metrics, a typical example is RUBER \citep{tao2018ruber} which combines a referenced metric and a reference-free metric. The referenced metric computes the cosine similarity of word embeddings between a generated response and a human reference, while a reference-free RNN-based scorer measures the relatedness between a generated response and its context. Based on it, BERT-RUBER \citep{ghazarian-etal-2019-better} replaces the RNN in RUBER with BERT \citep{devlin2019bert} to employ contextualized word embeddings as an improvement.

As for reference-free metrics, \citet{lan2020pone} proposed PONE based on BERT-RUBER to distinguish positive and negative samples from a training set. Besides, MAUDE \citep{sinha2020learning} adopts a similar training paradigm but with a different sampling strategy and uses the predicted score of a generated utterance given its context as the metric. Differently, FED \citep{mehri-eskenazi-2020-unsupervised} uses DialoGPT to compute the likelihood of manually designed follow-up utterances to measure multiple qualities of dialog without any supervision.

Our \textit{Gricean Scores} belong to reference-free metrics, and we additionally consider the measurement from the perspective of knowledge base. Another difference is that we utilize different pre-trained language models for different measurements, which are aligned well with their pre-training tasks, so that the gap between downstream inference and pre-training may not be large.

\section{Limitations and Future Work}
Our work has the following limitations. First, the performance of our knowledge selector is limited by the knowledge embeddings, e.g., TransE. Since the knowledge embeddings mainly focus on the graph structure instead of the semantic meanings of the entities or relations, the selector has a constrained feature space to learn how humans select follow-up question topics and focus. Second, we used a commonsense knowledge graph, Freebase. Although our framework could be adapted to any knowledge graph, the current use of Freebase is not tailored to the survey objective. A conversational survey may have survey objectives that require domain knowledge, such as medical information, a commonsense knowledge graph may not satisfy the need for domain knowledge. Third, the ecological validity of our evaluation is limited. In the future, we plan to integrate our model into a conversational survey and evaluate its effectiveness in information collection with real-world users.

\section{Conclusion}
We propose a knowledge-driven framework to address three challenges in follow-up question generation in conversational surveys, diverse and complex responses interpretation, high-quality question construction, and question evaluation. Our framework first selects an entity-relation pair from dialogue histories as question topic and focus, and then uses it to guide a GPT-based model to generate high-quality follow-up questions. To verify the effectiveness of the proposed framework, we collect a new dataset and propose a new set of reference-free evaluation metrics, \textit{Gricean Scores}. Extensive experimental results suggest that our framework outperforms competitive baseline models in both quantitative and qualitative experiments. 

\bibliography{custom}

\appendix

\section{Appendix}
\subsection{Implementation Details}
\label{appendex:implementation}
All the knowledge selection models use BERT$_{\text{large}}$ as the backbone encoder which is initialized with the pre-trained uncased weights. We use OpenKE\footnote{\url{http://openke.thunlp.org/}} to pre-train all knowledge embeddings. Models are implemented by Pytorch framework \citep{NEURIPS2019_bdbca288} and Huggingface transformers \citep{wolf-etal-2020-transformers}. We tune the parameters of each knowledge selection model with the following search space (bold indicate the choice for our final model according to the performance on the validation set): KG embeddings size: $\{100, 200, 300, \mathbf{400}, 500\}$, hidden states: $\{100, 200, 300, 400, 500, \mathbf{600}, 700\}$. The selection models are optimized by AdamW \citep{loshchilov2018decoupled} with the learning rate of $4e-5$ and the linear learning scheduler. A default setting trains for $50$ epochs, using a batch size of $20$. Early stopping is adopted if performance on the validation set doesn’t increase for consecutive $10$ epochs.

As for the generative model, we initialize it with DialoGPT$_{\text{large}}$ and also use AdamW with the linear learning scheduler to fine-tune models. The learning rate is set to $5e-5$. Similarly, we train the model for $5$ epochs with the batch size $1$ by default, and use early stopping to stop the training when the performance on the validation set doesn’t improve for consecutive $5$ epochs. During the generation, we use beam search with beam size $2$.

\subsection{Dataset Statistics}
\label{appendix:statistics}
The final dataset we collect consists of $10040$ dialogues, which we divide into $8032$ for train, $1004$ for validation, and $1004$ for the test, and we present a comparison between ours and related datasets in Table \ref{table:stat}. A total of $8165$ unique entities are mentioned in the question-answer pairs. On average, each question-answer pair mentioned $2.45$ unique entities. And for each mentioned entity, the average number of connected relations is $9.44$. More details are shown in Table~\ref{tab:my-table}. Regarding the annotated follow-up question for each question-answer pair, there are a total of $3917$ 'What' questions, $81$ 'How' questions, $341$ 'When' questions, $2619$ 'Who' questions, $12$ 'Why' questions, $806$ 'Which' questions, $716$ 'Where' Questions, and $1548$ closed-ended questions.

\begin{table*}[t]
\centering
\scalebox{0.9}{
\begin{tabular}{lccc}
\toprule \textbf{Datasets}      & \textbf{Size} & \textbf{Domain} & \textbf{Knowledge Base}        \\ \midrule
MHMC-IV \citep{su2018follow}   & 3.4k             & School Admission          & \xmark    \\
Interview Coaching \citep{su2019follow}      & 1.2k             & School Admission          & \cmark (ConceptNet) \\
FQG \citep{sb2020automatic}     & 1k             & Job Interviews          & \xmark \\ \midrule
\textsc{Ours}      &     10k         & General         &        \cmark(Freebase) \\ \bottomrule     
\end{tabular}}
\caption{Statistics of our dataset and other previous datasets for follow-up question generation.}
\label{table:stat}
\vspace{-0.2cm}
\end{table*}

\begin{table*}[ht]
\centering
\begin{tabular}{llll}
\hline
                           & Train  & Validation & Test  \\ \hline
Number of Dialogues        & 8032   & 1004       & 1004  \\
Number of Utterances       & 311298 & 38739      & 38002 \\
Number of Unique Entities  & 7429   & 1910       & 1920  \\
Avg. Entities per Dialogue & 2.44   & 2.45       & 2.48  \\ \hline
\end{tabular}
\caption{Dataset statistics of the Knowledge-Driven Follow-up Question Generation Task.}
\label{tab:my-table}
\end{table*}

\end{document}